\documentclass[journal]{IEEEtran}
\usepackage{microtype}

\usepackage{cite}

\ifCLASSINFOpdf
\else
\fi
\usepackage{amsmath}
\usepackage{amssymb}
\usepackage{algorithm}[H] 
\usepackage[noend]{algpseudocode}

\newbox\statebox
\newcommand{\myState}[1]{%
    \setbox\statebox=\vbox{#1}%
    \edef\thealgruleheight{\dimexpr \the\ht\statebox+1pt\relax}%
    \edef\thealgruledepth{\dimexpr \the\dp\statebox+1pt\relax}%
    \ifdim\thealgruleheight<.75\baselineskip
        \def\thealgruleheight{\dimexpr .75\baselineskip+1pt\relax}%
    \fi
    \ifdim\thealgruledepth<.25\baselineskip
        \def\thealgruledepth{\dimexpr .25\baselineskip+1pt\relax}%
    \fi
    \State #1%
    \def\thealgruleheight{\dimexpr .75\baselineskip+1pt\relax}%
    \def\thealgruledepth{\dimexpr .25\baselineskip+1pt\relax}%
}

\usepackage{array}
\usepackage{booktabs}

\ifCLASSOPTIONcompsoc
 \usepackage[caption=false,font=normalsize,labelfont=sf,textfont=sf]{subfig}
\else
 \usepackage[caption=false,font=footnotesize]{subfig}
\fi

\usepackage{threeparttable}

\usepackage{url}

\usepackage{tikz}
\usetikzlibrary{bayesnet}
\usetikzlibrary{arrows}

\renewcommand{\vec}[1]{{\boldsymbol{{#1}}}} 

\newcommand{\B}{\mathcal{B}}

\newcommand{\cC}{\mathcal{C}}
\newcommand{\cT}{\mathcal{T}}

\newcommand{\cB}{\mathcal{B}}

\newcommand{\card}[1]{\lvert #1 \rvert}
\newcommand{\abs}[1]{\lvert #1 \rvert}
\newcommand{\scC}{\card{\cC}}

\newcommand{\targetscC}{N_{\cC}^{target}}

\newcommand{\thresholdNov}{d_{\text{min}}}

\newcommand{\encoder}{\mathcal{E}}

\newcommand{\bdtaskspace}{\B_\cT}
\newcommand{\bdtask}{\vec b_{\cT}}
\newcommand{\bdprop}{\vec b_{\encoder}}

\newcommand{\mapelites}{MAP-Elites}
\newcommand{\cvtmapelites}{CVT-MAP-Elites}
\newcommand{\clusterelites}{Cluster-Elites}
\newcommand{\knn}{k\text{-NN}}

\newcommand{\vect}[1]{\boldsymbol{\mathbf{#1}}}

\algnewcommand\algorithmicforeach{\textbf{for each}}
\algdef{S}[FOR]{ForEach}[1]{\algorithmicforeach\ #1\ \algorithmicdo}

\newcommand{\obs}{\vec{sd} }

  \newcommand\indiv{indiv}

  \newcommand\perf{f}
  
  \newcommand{\bdpropindiv}{\bdprop^{ \indiv  }}

  \newcommand{\obsindiv}{\obs^{ \indiv  }}
  
  \newcommand\container{\cC}
\newcommand\perfindiv{\perf^{\indiv}}

\newcommand{\aurorapsat}{AURORA-CSC}
\newcommand{\auroravat}{AURORA-VAT}

\newcommand{\constantCSC}{K_{CSC}}
\newcommand{\constantVAT}{K_{VAT}}

\begin{document}
%
\title{Unsupervised Behaviour Discovery\\with Quality-Diversity Optimisation}
%
%
%

\author{Luca~Grillotti and
        Antoine~Cully
\thanks{All authors are members of the Adaptive and Intelligent Robotics Laboratory, Department of Computing, Imperial College London, United Kingdom (e-mail: \{luca.grillotti16; a.cully\}@imperial.ac.uk).}
}

%
%

\markboth{}%
{Shell \MakeLowercase{\textit{et al.}}: Bare Demo of IEEEtran.cls for IEEE Journals}
%



\maketitle

\begin{abstract}

Quality-Diversity algorithms refer to a class of evolutionary algorithms designed to find a collection of diverse and high-performing solutions to a given problem.
In robotics, such algorithms can be used for generating a collection of controllers covering most of the possible behaviours of a robot.
To do so, these algorithms associate a behavioural descriptor to each of these behaviours.
Each behavioural descriptor is used for estimating the novelty of one behaviour compared to the others.
In most existing algorithms, the behavioural descriptor needs to be hand-coded, thus requiring prior knowledge about the task to solve.
In this paper, we introduce: Autonomous Robots Realising their Abilities, an algorithm that uses a dimensionality reduction technique to automatically learn behavioural descriptors based on raw sensory data.
The performance of this algorithm is assessed on three robotic tasks in simulation.
The experimental results show that it performs similarly to traditional hand-coded approaches without the requirement to provide any hand-coded behavioural descriptor.
In the collection of diverse and high-performing solutions, it also manages to find behaviours that are novel with respect to more features than its hand-coded baselines.
Finally, we introduce a variant of the algorithm which is robust to the dimensionality of the behavioural descriptor space.
\end{abstract}


%
\IEEEpeerreviewmaketitle

%
%
%
%

\section{Introduction}

\IEEEPARstart{F}{ollowing} the recent improvements in the robotics field, robots can now complete tasks that are too complex or too risky for humans, including inspecting the damages on nuclear plants \cite{nagatani2013emergency}, extinguishing flames \cite{peskoe2019paris} and exploring other planets \cite{sanderson2010mars}. 
In such critical missions, robots should exhibit some versatility and resilience, especially if they fall into an unforeseen situation. 
From an engineering point of view, it is technically challenging to anticipate every possible situation a robot could have to face.
One solution to this challenge is to enable robots to discover their abilities autonomously.
With such autonomous learning, engineers would not be required to hand-craft behaviours and controllers for all the different situations the robot may face.
Having robots that discover autonomously their abilities would not only improve their autonomy and resilience, but also reduce the number of technical challenges required to program them.

This problem of learning diverse behaviours for a robotic system can be addressed via Quality-Diversity (QD) algorithms. 
QD algorithms are a recently introduced class of evolutionary algorithms \cite{Pugh2015, Pugh2016} that aim to find not only a single solution to an optimisation problem, but instead generate a large collection of diverse and high-performing solutions. 
In robotics, QD algorithms have been particularly used for generating collections of diverse behaviours, also called behavioural repertoires~\cite{Cully2014}
Such collections have been used to solve diverse tasks such as legged locomotion~\cite{duarte2017evolution, Lehman2011}, maze navigation~\cite{gravina2018quality} and ball throwing~\cite{kim2017learning}.
In real-world environments, the diversity of behaviours present in such collections can be exploited to efficiently recover from damage ~\cite{Cully2014, chatzilygeroudis2018reset}.

\begin{figure}[t]
\centering
\includegraphics[width=0.45\textwidth]{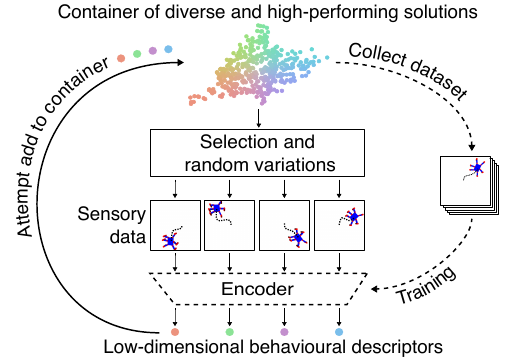}
\caption{Inner working of AURORA. 
AURORA alternates between two phases: a Quality-Diversity phase (plain arrows) and an encoder update phase (dashed arrows).
The main components introduced in AURORA are presented using dashed arrows and contours. 
In the Quality-Diversity phase, the first step selects parent solutions from the container. 
Those parent solutions are then copied and undergo random variations (e.g. mutations, cross-overs).
After that, we evaluate their fitness and the sensory data that they collect.
An encoder then maps that sensory data into low-dimensional behavioural descriptors.
If the novelty of their behavioural descriptor and the value of their fitness are high enough, they can be added to the container.
During the encoder update phase, the encoder is trained with the sensory data collected by the individuals in the container.
Each time the encoder is updated, it is used to re-compute the behavioural descriptors for all the individuals in the container.
}
\label{fig:aurora_procedure}
\end{figure}

QD algorithms build these collections of behaviours by adding them in a container only if they are novel or high-performing enough. While the performance of the behaviours is computed via a fitness function, their novelty is computed by associating a Behavioural Descriptor (BD) to each behaviour. This BD is a numerical vector that characterises the overall behaviour. For instance, the BD can be the final location of the robot after executing a controller for a few seconds~\cite{Cully2014}. The novelty of a behaviour is defined as the minimal or average distance between its BD and the BDs of the behaviours already present in the container.
In most QD algorithms, the definition of the BD is low-dimensional and hand-coded.
Nevertheless, relying on hand-coded BDs can be an issue for two reasons.
First, the choice of an appropriate BD is not straightforward; and several relevant BD definitions may lead to various performance for the QD algorithm \cite{bossens2020learning}.
That means some prior knowledge about the task is required to define an appropriate BD.
Such reliance on prior knowledge is the basis of potential bottlenecks in the design process of a robot, as the appropriate BD definition depends on many parameters: robot morphology, task to solve, environment, damages... 
In some situations, the behaviours are non-trivial to characterise using a low-dimensional BD, such as when the BD must be defined based on a robot picture, or based on the full trajectory of an object pushed by a robot.
Second, hand-coded BDs may constrain the diversity of the container.
For example, if the hand-coded BD corresponds to the end position $(x_T,y_T)$  of an object, then the QD algorithm intends to learn similar trajectories leading to diverse positions.
However, in that case, the QD algorithm does not aim at learning diverse trajectories.
In other words, QD algorithms only consider diversity with respect to hand-coded BDs.
The diversity obtained via QD algorithms is thus biased and possibly sub-optimal.

To avoid those issues, we present \textit{AUtonomous RObots that Realise their Abilities} (AURORA), a QD algorithm that autonomously learns how to define appropriate BDs based on some raw sensory data collected by the robot. 
The main difference between AURORA and standard QD algorithms is that standard QD algorithms rely on BDs which are hand-coded and low-dimensional \cite{Cully2018, tarapore2016differentencodingsinfluence}; whereas AURORA directly uses raw sensory data which is possibly high-dimensional without any prior from the users.
More specifically, AURORA uses a dimensionality reduction algorithm to encode the sensory data into unsupervised low-dimensional BDs. 
That reduction algorithm is iteratively trained and refined with the solutions from the container.

We evaluate our approach on three QD simulated robotic tasks, with various types of high-dimensional sensory data, such as colour pictures from a camera and complete trajectories of an object.
In those tasks, AURORA is compared to several QD baselines, which are provided with hand-coded BDs.
Our results show that AURORA can generate collections of diverse and high-performing behaviours that are similar to standard QD algorithms despite the lack of prior from the users. 
Furthermore, our analysis reveals a correlation between the unsupervised BDs learned by AURORA and the ground truth states of the robot.
Finally, we show that the absence of priors from the users enables AURORA to discover a wider range of novel behaviours that are not constrained by the definition of the BD.

While we originally introduced the AURORA algorithm in a conference paper~\cite{Cully2019}, we present in this paper an extended analysis considering exclusively new experiments, as well as additional improvements in the context of high-dimensional BD spaces.
More specifically, the main contributions of this paper are: 

\begin{enumerate}
    \item We evaluate AURORA in three tasks of varying difficulty, where different types of sensory data are collected. We show that AURORA finds collections of solutions which are not only diverse, but also high-performing (in similar previous works \cite{Cully2019, Paolo2019}, the performance was not considered).
    \item We provide an extended analysis showing that AURORA can find a higher diversity of behaviors than hand-coded QD algorithms. 
    \item We introduce a new variant of AURORA called "AURORA with Container Size Control" (\aurorapsat{}), which improves the robustness of the algorithm in presence of high-dimensional BD spaces.
\end{enumerate}

\section{Background}

\subsection{Quality-Diversity Algorithms}

\label{sec:background:qd_algo}

Our problem consists of finding a collection of individuals (solutions) that are both diverse and high-performing for a given task. 
Quality-Diversity (QD) algorithms are evolutionary algorithms that have been designed for this purpose \cite{Pugh2015, Pugh2016}.
In QD algorithms, the collection of individuals is stored in a \emph{container}; and each individual's behaviour is characterised via a feature vector: the \emph{Behavioural Descriptor} (BD).
For example, the BD of a moving robot can correspond to its final position.
QD algorithms optimise the diversity of the container with respect to these BDs.
More precisely, QD algorithms aim at finding a container of individuals maximising the coverage in the BD space.
At the same time, QD algorithms also locally optimise the performance of each individual.
Such performance is measured via a \emph{fitness function}. 
In summary, QD algorithms intend to (1) maximise the diversity of the individuals' BDs, and (2) locally maximise each individual's fitness.

In standard QD algorithms, like \mapelites{}~\cite{Mouret2015}, each iteration consists of four steps~\cite{Cully2018}.
First, some individuals are selected from the container according to a selection strategy.
Then, those selected individuals are copied and undergo random variations, such as cross-overs and mutations.
After that, their fitness and BD are evaluated.
In the end, the individuals are added to the container if they fulfil some conditions specific to the container.
Those container conditions for adding new individuals are designed to progressively improve the overall diversity and fitness of the individuals in the container.
Such conditions depend on the type of container used~\cite{Cully2018}.
There are two main categories of containers: unstructured containers and grid-based containers.
The unstructured container \cite{Cully2018, Lehman2011} uses a distance threshold $\thresholdNov$ in the BD space to decide when solutions should be added to the container. Conversely, grid-based containers rely on a discretisation of the behaviour space to preserve diversity \cite{Mouret2015, Cully2014}.

The selection of the individual at the beginning of each iteration is governed by a selection operator.
For example, a standard selection strategy consists of selecting the individuals from the container uniformly at random, like in \mapelites{} \cite{Mouret2015}. 
Other types of selectors may be used to improve the performance of the QD optimisation.
For example, the novelty-based selection strategy \cite{Pugh2015} aims at improving the diversity of the QD container in priority.
To do so, novelty-based selectors choose with a higher probability the individuals having the highest novelty score.

\subsection{Learning Diverse Behaviours from High-Dimensional Spaces}

\label{subsec:highdim}

The BD can correspond to any type of data, including raw high-dimensional data.
For example, full pictures taken from a camera can be used as a BD \cite{maestre2015bootstrapping}.
However, the computational complexity of each QD iteration rises significantly as the number of BD dimensions increases. 
For example, the state-of-the-art QD algorithm \mapelites{} \cite{Mouret2015} requires the behaviour space to be discretised into a grid.
However, the number of grid bins to manage increases exponentially with the number of dimensions of the behaviour space.

The \cvtmapelites{} algorithm \cite{Vassiliades2018a} addresses that issue by generating directly a specified number of bins $k$ in the behaviour space.
To do so, \cvtmapelites{} constructs a Centroidal Voronoi Tessellation (CVT) grid according to a pre-defined dataset of possible behaviours.
The CVT grid is constructed using the $k$-means algorithm \cite{macqueen1967some} applied to that dataset.
The $k$ resulting centroids define the bins of the CVT grid.
If there is no pre-defined dataset available, \cvtmapelites{} generates the dataset based on a uniform sampling of the BD space.
That uniform sampling is performed according to the bounds of each dimension. 
However, the BDs that are achievable may be only a small subset of the full BD space.
For example, the set of possible pictures taken from a camera is a tiny subset of all possible images (containing all combinations of pixel colours).
In that case, the achievable BDs are all contained in a limited number of bins, which affects negatively the performance of \cvtmapelites{} \cite{Cully2019}.

Similarly, unstructured containers always rely on the $k$-Nearest Neighbours ($\knn{}$) algorithm to compute the novelty score of each individual. 
If $\cB$ represents the BD space, and $\cC$ represents the container, then the complexity of a $\knn{}$ applied to one individual is $O\left( \dim{\cB}\log \abs{\cC}  \right)$ in average \cite{bentley1975multidimensional}.
Furthermore, in such QD algorithms, a $\knn{}$ search is required for all individuals at every QD iteration \cite{Cully2018} in order to update all the novelty scores.
Consequently, the complexity resulting from all the $k$-NN searches throughout the QD optimisation is $O(I \abs{\cC}  \dim{\cB} \log \abs{\cC})$, where $I$ represent the number of iterations of the QD algorithm.
Hence with unstructured containers, the overall QD optimisation may become too computationally expensive if the dimension of $\cB$ increases significantly.

Moreover, in QD algorithms, the novelty of an individual is calculated using the euclidian distance.
But the euclidian distance is not adapted to every task.
For example, it is not always relevant to measure the difference between two images or two videos by calculating the direct euclidian distance between them.
It would be more relevant, in this case, to have a mechanism extracting the relevant features from those images or videos. 
Those extracted features can then be used for estimating distances between individuals.
For instance, in the work from Maestre et al. \cite{maestre2015bootstrapping}, the BD is calculated by using a hand-crafted method for encoding images.
In the end, the BD of each individual corresponds to a hash of its associated image.
The distance between two individuals is then computed by using a specific distance for comparing hashes: the Hamming distance \cite{hamming1950error}.

In this work, we propose AURORA, a QD algorithm that learns low-dimensional BDs in an unsupervised manner using dimensionality reduction algorithms.
The distance between individuals corresponds then to the distance between those unsupervised BDs.

\section{Related Work}

\subsection{Learning Appropriate Behavioural Descriptors}

\label{subseq:AURORA}

Several works in the Quality-Diversity (QD) literature study the problem of defining automatically appropriate Behavioural Descriptors (BDs).
For instance, meta-learning has been used to find an optimal BD definition to improve the success rate in multiple tasks \cite{meyerson2016learning}, or to improve the behavioural diversity in multiple environments \cite{bossens2020learning}.
Instead of relying on a meta-objective, AURORA learns the BDs in an unsupervised manner from raw sensory data.

Another common approach is to rely on a pre-defined dataset to learn the BDs.
For example, in hierarchical behavioural repertoires \cite{Cully2018a}, before starting the QD optimisation, an Auto-Encoder (AE) is trained to learn unsupervised descriptors of the MNIST dataset \cite{lecun1998gradient}. 
After that, during the QD optimisation, the previously trained encoder is used to generate the descriptors for digits written by robotic arms.
In a similar manner, IMGEP-UGL \cite{Pere2018} starts by learning an unsupervised goal space from a pre-defined dataset.
That goal space is then used for sampling goals that the agent learns to accomplish.
As another example, Innovation Engines \cite{nguyen2015innovation} pre-train an "AlexNet" \cite{krizhevsky2012imagenetalexnet} on the ImageNet dataset \cite{deng2009imagenet} and use the label predictions of the network as BDs.
This way, by looking for new BDs, Innovation Engines discover new types of images.
In all those works, a pre-defined dataset is necessary for learning an appropriate BD space before the QD optimisation.

Conversely, AURORA builds its dataset automatically and updates regularly its encoder.
Similarly, Deep Learning Novelty Explorer (DeLeNoX) \cite{Liapis2013} and Task Agnostic eXploration of Outcome spaces through Novelty and Surprise (TAXONS) \cite{Paolo2019} alternate between a QD optimisation phase, and a "training phase" \cite{Liapis2013}.
In DeLeNoX, the Auto-Encoder (AE) is always reset before being trained; whereas TAXONS and AURORA learn the BD space in an online manner.
Indeed, in TAXONS and AURORA, the AE is never reset before being updated: the learning of the BD space is based on the knowledge previously accumulated by the AE.
Contrary to AURORA, DeLeNoX and TAXONS are based on the Novelty Search (NS) framework \cite{Lehman2011}.
As such, they do not present any mechanism to maximise the fitness of the individuals.
In those NS-based algorithms, a constant number of individuals is added to the container at each iteration, without removal nor replacement: the container size grows linearly with the number of iterations.

\subsection{Container Size Management in Unbounded Spaces}

\label{sec:unbounded}

Having too many individuals in the container may make the QD algorithm more computationally expensive.
In unbounded spaces, previous works have studied ways to prevent the container size from being too elevated.

Similarly to AURORA, Go-Explore \cite{ecoffet2021firstreturnthenexplore} adjusts its encoding to control the container size.
Go-Explore downscales in-game images of the Atari environment, and uses the resulting downscaled images as BDs.
The parameters of this downscaling (e.g. the final resolution, the pixel depth) are automatically adjusted to obtain the desired number of different downscaled images.

In order to constrain the container size in unknown BD spaces, Vassiliades et al. \cite{Vassiliades2017} introduce expansive variants for \mapelites{} and \cvtmapelites{}, as well as another approach called \clusterelites{}.
Those variants progressively adapt the volume of the grid bins to the shape of the explored BD space.
\clusterelites{} and the expansive variant of \cvtmapelites{} rely on the computationally expensive operation of constructing a CVT periodically.
To avoid those issues, AURORA does not rely on a grid-based container, and uses an unstructured container with an adaptive distance threshold.

\section{Autonomous Behaviour Discovery from Unsupervised Behavioural Descriptors}

The aim pursued by this work is to enable robots to autonomously discover their abilities. 
For this, we propose AURORA, a Quality-Diversity (QD) algorithm for autonomously learning unsupervised behavioural descriptors in an online manner.
A high-level representation of the inner working of AURORA is depicted in Figure~\ref{fig:aurora_procedure}.

\subsection{Learning Behavioural Descriptors from Sensory Data in an Online Manner}

\label{sec:discovering_unsupervised_behaviours}

AURORA alternates between two phases: a QD phase, and an encoder update phase.
During the QD phase, AURORA discovers new behaviours by filling the space of unsupervised Behavioural Descriptors (BDs).
In the encoder update phase, this unsupervised BD space is refined based on the behaviours that have been discovered.
The encoder update thus provides a refined BD space that will be filled in the following QD phase.

\subsubsection{Quality-Diversity Phase}

AURORA follows the same structure as any standard QD algorithm.
At each iteration, a batch of individuals is selected from the container, copied, mutated, evaluated, and finally potentially added to the container.
Nevertheless, our problem setting differs from standard QD optimisation tasks in the way individuals are evaluated.
In standard QD tasks, the evaluation of an individual returns its BD, which is hand-coded and low-dimensional.
In AURORA, the evaluation of an individual returns some sensory data collected by that individual during its evaluation; that sensory data is likely to be high-dimensional.

To process such high-dimensional sensory data, AURORA uses a dimensionality reduction algorithm.
Any dimensionality reduction technique can be used, such as a Principal Component Analysis or a deep convolutional Auto-Encoder (AE) \cite{masci2011stacked}.
The dimensionality reduction algorithm, referred to as \emph{encoder}, projects the sensory data into a low-dimensional latent space.
The resulting projections are used as BDs for the individuals, since they encode the features present in the sensory data.
As explained in section \ref{subsec:highdim}, using the high-dimensional sensory data as BD is feasible in certain cases, but computationally expensive.
AURORA avoids that issue by training its encoder to learn low-dimensional BDs in an unsupervised manner.

\subsubsection{Encoder Update Phase}

\label{sec:encoder_updafte_phase}

This encoder needs to be updated regularly to take into account the sensory data collected by new individuals.
To train the encoder, the sensory data of all the individuals in the container are collected and used as a training dataset.

However, each time the encoder is trained, the entire latent space changes, leading to a modification of all the BDs.
Consequently, after each encoder training, all the BDs from the container are updated based on the up-to-date encoder.


During the first iterations, QD algorithms mainly encounter novel solutions which are likely to exhibit unknown sensory data.
In that case, the encoder needs to be updated frequently.
This way, the encoder keeps on capturing the relevant features in the varying sensory data.
On the contrary, at the end of the evolutionary process, the content of the container varies slowly, and the encoder does not need to be trained as frequently.
In AURORA, the interval between two encoder updates is increased linearly.
For example, supposing that the first encoder training is performed at the $10^{\text{th}}$ QD iteration, the encoder updates will occur at the following iterations: 10, 30, 60, 100, 150...


The BDs are learned in an online manner by the encoder.
The encoder is never reset during an experiment: it is initialised once (before the first QD iteration), and it is never reset afterwards.
In other words, its parameters are kept fixed between two encoder updates and each encoder update fine-tunes the structure of the BD space.

\label{sec:learning-low-from-high}

\newcommand{\mi}[2]{I\left( #1 , #2 \right)}

\subsection{Managing the Container Size in Unpredictable and Unbounded Descriptor Spaces}

\label{sec:methods:managing_container_size}

The structure of the BD space learned by the encoder depends on the task and the dimensionality reduction algorithm used.
In the case of standard auto-encoders, the distribution of BDs in the latent space is likely to change between two runs of the same experiment.
Hence, the magnitude of the distances between BDs may vary significantly between experimental runs, and even between the updates of the encoder.
In this situation, it is difficult to estimate when an individual is novel enough to be added to the container.
Consequently, AURORA needs to adapt to the varying BD distribution of the container.

As in standard QD algorithms, an individual is added to the container if the distance of its BD to its nearest neighbour is higher than a distance threshold $\thresholdNov$.
In QD algorithms, that threshold $\thresholdNov$ is usually fixed and defined by the user \cite{Cully2018, Cully2018a}. 
However, as explained before, the latent space may vary between several runs of the same experiment.
Thus, the relevant value for the distance threshold $\thresholdNov$ may change between several runs.
And the value of this distance threshold $\thresholdNov$ significantly impacts the overall performance of the algorithm.
If the threshold $\thresholdNov$ is too high, then it will be difficult for the QD algorithm to find individuals which are novel enough.
Conversely, if the threshold $\thresholdNov$ is too low, then too many individuals may be considered novel enough to be added to the container, making the QD algorithm more computationally expensive.
In AURORA, a desired container size $\targetscC$ is defined.
Our purpose is to control the number of individuals in the container (referred to as \emph{container size}) $\scC$ so that it tends to $\targetscC$.

In this work, we study two methods for adapting the distance threshold $\thresholdNov$ so that the container size tends to $\targetscC$: the Volume Adaptive Threshold (VAT) and the Container Size Control (CSC).

\subsubsection{Volume Adaptive Threshold}

Used by Cully \cite{Cully2019}, the distance threshold $\thresholdNov$ is adapted to the volume of the smallest sphere containing all the unsupervised BDs.
More precisely, $\thresholdNov$ is chosen such that: if the container has $\targetscC$ individuals, then the sphere would be fully covered by the individuals in the container.
From a mathematical point of view, $\thresholdNov$ is chosen to verify the following equality: 
\begin{align}
  \label{eq:equality-constraint-vat}
    V_{sph}^{\cC} = \constantVAT\times \targetscC \times V_{sph}^{indiv}
\end{align}
where $V_{sph}^{\cC}$ is the volume of the smallest sphere containing all the individuals in the container; $V_{sph}^{indiv}$ is the volume taken by an individual, assuming its radius is $\frac{\thresholdNov}{2}$; and $\constantVAT$ is a constant adjusted to ultimately obtain $\targetscC$ individuals in the container.
In summary, after each encoder update, the VAT method changes $\thresholdNov$ in the following way:
\begin{align}
  \label{eq:update-vat-l}
    \thresholdNov \leftarrow \dfrac{\text{maximal BD distance in }\cC}{\left(\constantVAT \times \targetscC\right)^{\frac{1}{n}}}
\end{align}
where $n$ represents the dimensionality of the BD space.

However, equality~\ref{eq:equality-constraint-vat} assumes that all the descriptors are uniformly distributed in the sphere envelope. 
This approximation leads to estimation errors on the appropriate $\thresholdNov$ (in update rule~\ref{eq:update-vat-l}).
These errors are not significant in two dimensions but tend to increase exponentially with the dimension of the latent space, due to the curse of dimensionality \cite{bellman1961adaptive}.

Thereafter, we call \emph{\auroravat{} (AURORA with Volume Adaptive Threshold)}, the variant of AURORA that uses this distance threshold adaptation.

\subsubsection{Container Size Control}

\begin{algorithm*}[t]
  \small
  \caption{\aurorapsat{} (number of iterations $I$; encoder $\encoder$; target container size $\targetscC$, container update period $T_\cC$)}
  \label{algo:aurora}
  \newcommand\algorithmicitemindent{\hspace*{\algorithmicindent}\hspace*{\algorithmicindent}}

  \begin{algorithmic}[1]
    \State{$\container \gets \emptyset$}  \Comment{\textit{Start with empty unstructured container.}}
    \State{Initialise $\thresholdNov$  \Comment{\textit{Initialise minimum-distance threshold for container $\container$.}}}
    \For{$iter$ = 1 $\rightarrow I$} \Comment{\textit{Perform $I$ iterations of the main loop.}}
    \If{first iteration} \Comment{\textit{Generate random parents and offspring at first iteration.}}
      \State{$\vect{pop_{parents}} \gets $ random()}
      \State{$\vect{pop_{offspring}} \gets $ random()}
    \Else
        \State{$\vect{pop_{parents}} \gets $ selection($\container$) \Comment{\textit{Choose parents by selecting from the container.}}}
        \State{$\vect{pop_{offspring}} \gets $ random\_variation($\vect{pop_{parents}}$) \Comment{\textit{Generate offspring via mutations and cross-overs on parents.}}}
    \EndIf
    \ForEach{$indiv$ in $\vect{pop_{offspring}}$} \Comment{\textit{Evaluation of individuals in offspring population}}
    \State{$\{\obsindiv,  \perfindiv\}  \gets$ evaluate($indiv$)  \Comment{\textit{Compute sensory data $\obs$ and performance $\perf$ of offspring individual.}} }
    \State{$\bdpropindiv \gets\encoder(\obsindiv)$} \Comment{\textit{Encode the sensory data $\obs$ into a low-dimensional behavioural descriptor (BD) $\bdprop$.}}
    \State{$\container \gets$ try\_add\_to\_container($indiv$, $\container$)} \Comment{\textit{Attempt to add individual to $\container$, depending on its performance} $\perfindiv$ \textit{and BD} $\bdpropindiv$.}
    \EndFor
    \If{encoder update expected at iteration $iter$} \Comment{\textit{At iterations 10, 30, 60, 100... (spacing increased linearly).}}
    \State{$\encoder \gets$ train($\encoder$, $\container$) \Comment{\textit{Update encoder $\encoder$ with the sensory data from all individuals present in the container $\container$.}}}
    \State{$\container \gets$ update\_descriptors\_in\_container($\container$, $\encoder$)} \Comment{\textit{Re-compute descriptors $\bdprop$ for individuals in container with updated encoder $\encoder$.}} 
    \EndIf
    \State{$\thresholdNov \gets$ update\_threshold\_distance($\thresholdNov$, $\scC$, $\targetscC$) \Comment{\textit{Update $\thresholdNov$ via rule~\ref{eq:update-l}, in the case of \aurorapsat{}}}}\label{algo:aurora:update_threshold}
    \If{$iter$ multiple of $T_\cC$} \Comment{\textit{Periodic update of the container.}}
        \State{$\container \gets$ update\_container($\container$, $\thresholdNov$)} \Comment{\textit{Remove and re-add all individuals to container with new distance threshold.}} \label{algo:aurora:update_container}  

    \EndIf
      \EndFor\label{auroraendfor}
      \State{\Return container $\container$}
  \end{algorithmic}
\end{algorithm*}

\label{subseq:adaptive}

To avoid the estimation errors occurring in \auroravat{}, we propose another variant that uses a proportional control loop to control the container size $\scC$.
More precisely, the value of $\thresholdNov$ is corrected at each iteration to keep  $\scC$ around the desired container size $\targetscC$:
\begin{align}
  \label{eq:update-l}
  \thresholdNov \leftarrow \thresholdNov \times \left( 1 + \constantCSC \left( \scC  -  \targetscC \right) \right)
\end{align}
where $\constantCSC$ represents the gain of the proportional control.
Update rule~\ref{eq:update-l} reaches a fixed point for $\thresholdNov$ when the container size $\scC$ is equal to the targeted size $\targetscC$.
When $\scC$ is inferior to $\targetscC$, the value of the threshold novelty $\thresholdNov$ decreases, making it easier for new individuals to be added to the container.
On the contrary, when $\scC$ is superior to $\targetscC$, there are more individuals than expected in the container.
As a consequence, the threshold novelty $\thresholdNov$ increases, which makes it more difficult for new individuals to be added to the container. 
In the end, by adapting the value of $\thresholdNov$, the size of the container $\scC$ can be kept around the desired container size $\targetscC$.

The resulting algorithm corresponds to a variant of AURORA in which the distance threshold $\thresholdNov$ adapts to the size of the container.
We call that variant \emph{\aurorapsat{} (AURORA with Container Size Control)}. 

\bigskip

Nevertheless, adapting $\thresholdNov$ only affects the condition for adding new individuals to the container; whereas the individuals that are already in the container are not affected by a change of $\thresholdNov$.
To take into account the modifications of $\thresholdNov$, a container update needs to be performed. 
The container update consists of taking all the individuals out of the container, and then re-adding them to it with the up-to-date value of $\thresholdNov$.
The conditions for performing a container update are different between \auroravat{} and \aurorapsat{}.
In \auroravat{}, a container update is performed only after an encoder update.
This way, the estimation of the sphere envelope volume is adapted to the new BD space generated by the encoder.
In \aurorapsat{}, the container is periodically updated with a period $T_\cC$, manually defined as a hyperparameter.
This prevents the proportional control from diverging.

A pseudo-code for \aurorapsat{} is provided in Algorithm~\ref{algo:aurora}. 
The only difference with \auroravat{} is that: in \auroravat{}, the updates of $\thresholdNov$ and of the container $\cC$ occur only after an encoder update.

\section{Experimental Setup}

\subsection{Variants and Baselines}

In our experiments, we compare AURORA against variants using unsupervised BDs and variants using hand-coded BDs.
All the variants under study are described in Table~\ref{table:variants}.

\begin{table*}[t]
    \centering
\caption[Summary of the variants compared]{
    Quality-Diversity Variants under Study
    }
    \label{table:variants}

    \begin{threeparttable}
    \begin{tabular}{l l l l l l}
    \toprule
        Type of Behavioural Descriptor & Variant Name  & Distance Threshold Adaptation & Selection Operator  \\
        \midrule
            	\addlinespace

	Unsupervised (of size $n$) & \aurorapsat{}-uniform-$n$\tnote{1}   &     Container Size Control & Uniform \\
	& \aurorapsat{}-novelty-$n$  &     Container Size Control & Novelty-based \\
	& \aurorapsat{}-surprise-$n$ &   Container Size Control  & Surprise-based \\
    	
	& \auroravat{}-uniform-$n$\tnote{1}  & Volume Adaptive Threshold & Uniform \\
	& \auroravat{}-novelty-$n$ & Volume Adaptive Threshold & Novelty-based \\
	
    	& TAXONS-$n$ \cite{Paolo2019}   &     None  & Based on Novelty and Surprise \\

    	\addlinespace

    Hand-coded & HC-CSC-uniform\tnote{2}  &      Container Size Control & Uniform \\

        & Random Search  &      Container Size Control & No Selection \\
        & Novelty Search (NS) \cite{Lehman2011}  &      None & Novelty-based \\

    \bottomrule
    	
    \end{tabular}
    
    \begin{tablenotes}
         \item[1] CSC and VAT correspond respectively to the \textit{Container Size Control} and the \textit{Volume Adaptive Threshold} techniques. 
         \item[2] HC refers to \textit{Hand-Coded}.
     \end{tablenotes}
    \end{threeparttable}

\end{table*}

\subsubsection{Variants using Unsupervised Behavioural Descriptors}

We intend to show that some selector operators are beneficial to the performance of AURORA.
For that purpose, we tested the following selection operators:
\begin{itemize}
    \item Uniform selector: selecting individuals at random in the container.
    \item Novelty-based selector: selecting with probability proportional to the novelty score \cite{Pugh2015} (see section~\ref{sec:background:qd_algo}).
    \item Surprise-based selector: selecting with higher probability the individuals presenting the highest surprise score.  The surprise score refers to the reconstruction error from the encoder and has been introduced by Gaier et al. \cite{gaier2019quality} and used in the selection process of TAXONS \cite{Paolo2019}.

\end{itemize}

All these selectors were tested in conjunction with the Container Size Control (CSC) technique. 
To compare the CSC and Volume Adaptive Threshold (VAT) methods, we also consider \auroravat{} combined with a uniform selector (see Table~\ref{table:variants}).

Additionally, we evaluate the performance of TAXONS which also learns Behavioural Descriptors (BD) via an encoding.
Contrary to AURORA, TAXONS has a population-based selection.
Instead of selecting individuals from the container, TAXONS evolves a population separated from the container.
The way TAXONS manages its container is also substantially different.
A constant number of individuals $Q$ (the most novel or the most surprising) is added to the container at each generation.
Consequently, in the case of TAXONS, the number of individuals in the container increases linearly with the number of generations.

\subsubsection{Baselines using Hand-Coded Behavioural Descriptors}

To evaluate the minimal and maximal achievable performance of our approach, we compare AURORA to several omniscient QD algorithms, which use directly the hand-coded BD $\bdtask$, instead of the raw sensory data $\obs$.

\textit{HC-CSC-uniform:} A standard QD algorithm using an unstructured archive and a uniform selector (similarly to AURORA).
Furthermore, for a fair comparison with AURORA-CSC, this baseline also relies on the CSC technique to adapt the distance threshold $\thresholdNov$ used by the unstructured archive.

\textit{Novelty Search \cite{Lehman2011} (NS):} NS provides an upper bound on the supposedly achievable performance of TAXONS.
Contrary to TAXONS, NS uses directly the hand-coded BD; and NS presents some elitism absent from TAXONS (TAXONS systematically discards most of the parents at each iteration).
Our implementation is based on the description of the NS algorithm by Doncieux et al. \cite{Doncieux2020}.

\textit{Random Search:} Finally, to show the benefits of selecting individuals from the container in our experiments, we compare AURORA to a QD algorithm that does not select any individual from the container.
In that variant, all the new genotypes are sampled uniformly at random from the genotype space.
This random search variant is used to evaluate the difficulty of exploration in the BD space.

\begin{table*}[t]
    \centering
\caption[Experimental Tasks]{
    Experimental Tasks
    }
\begin{threeparttable}
    \begin{tabular}{l l l l l}
    \toprule
        Name Task & Sensory Data $\obs$ & Hand-coded BDs\tnote{1} $\bdtask$ & Hand-coded BD space\tnote{2} & Fitness $f$ \\
        \midrule
        Maze & $64\times 64$ image of the top of the maze & Final robot position & $\left[0, 600\right] \times \left[0, 600\right]$ & $-\sum_{t=1}^{T} \| \vec u_t \|^2$ \\
        Hexapod & $64\times 64$ top-down picture from the vertical of the origin & Final hexapod position & $\left[ -1,1 \right] \times \left[ -1.2, 1.2 \right]$  & $- |\alpha^{\indiv} - \alpha^{\text{circular}}|$ \\
        Air-Hockey & 50 successive positions of the puck $(x_t,y_t)_{t\in\left[1..T\right]}$ & Final puck position & $\left[-1, 1\right] \times \left[-1, 1\right]$ &  $-\sum_{t=1}^{T} \| \vec v^{\text{joints}}_t \|^2 $ \\
    \bottomrule
    \end{tabular}

    \begin{tablenotes}
         \item[1] BD refers to \textit{Behavioural Descriptor}.
         \item[2] The hand-coded BD space $\bdtaskspace$ is used to evaluate the coverage performance (see section~\ref{subsubsec:metrics-coverage}). That is why we consider bounded definitions for $\bdtaskspace$.
     \end{tablenotes}
    \end{threeparttable}

    \label{tab:variants_maze}
\end{table*}

\subsection{Tasks Under Study}

\label{sec:tasks_description}

The objective of our experiments is to assert the capabilities of AURORA to find a container of diverse high-performing behaviours for an agent.
In all the tasks under study, the raw sensory data collected by the agent $\obs$ always contains all the information related to the hand-coded BD $\bdtask$.
In other words, in each task, $\bdtask$ can be deduced from $\obs$.
Nevertheless, in general, $\bdtask$ contains strictly less information than $\obs$.
We will study three different QD tasks with various difficulties and varying levels of discrepancy between $\bdtask$ and $\obs$:
\begin{itemize}
    \item The \textit{Maze} task is almost identical to the one implemented by Paolo et al. \cite{Paolo2019}.
    The only difference is that we considered a non-constant fitness function, to stay consistent with our other tasks.
    In that task, there is almost no discrepancy between $\bdtask$ and $\obs$.
    \item The \textit{Hexapod} task is more challenging than the Maze task, in a sense that it requires more QD iterations before reaching a decent coverage.
    In that task, there is a low discrepancy between $\bdtask$ and $\obs$.
    \item The \textit{Air-hockey} task requires fewer iterations than the previous ones to achieve a full coverage of the BD space.
    There is however a high discrepancy between $\bdtask$ and $\obs$.
\end{itemize}

\begin{figure}
  \centering
\includegraphics[width=0.40\textwidth]{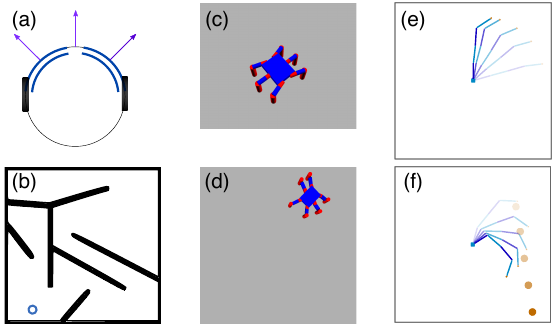}
  
\caption{Illustrations of the three tasks under study. 
(a) Robot used in the Maze task, with its two wheels, its three laser sensors (represented by arrows) and its two contact sensors (represented by double arc circles). 
(b) Topology of the maze; the blue circle represents the starting position of the robot.
(c) Close view of the hexapod robot in simulation.
(d) Example of sensory data in the Hexapod task: picture taken from a camera placed at the vertical of the origin.
(e) First phase of the Air-Hockey task: the robotic arm moves to a position.
(f) Second phase of the Air-Hockey task: a ball has appeared, and the robotic arm moves to another position.
}

 \label{fig:all_experiments}
\end{figure}

A summarised description of the tasks is provided in Table~\ref{tab:variants_maze} and illustrations are depicted in Figure~\ref{fig:all_experiments}.

\subsubsection{Maze}

In the \textit{Maze} task, a simulated circular robot is considered in a two-dimensional maze of size (600 units, 600 units). 
That robot possesses three laser sensors, placed respectively at angles of $-\frac{\pi}{4}$, $0$ and $+\frac{\pi}{4}$ radians compared to the directional axis of the robot. 
Those laser sensors have a maximum range of $100$ units. 
Also, the robot has two contact sensors. 
Each contact sensor covers a quarter of the robot perimeter; they cover respectively the top-left and top-right perimeter of the robot (Fig.~\ref{fig:all_experiments}). 
The environment is based on the experiment from Lehman et al. \cite{Lehman2011}; and the task is a replication of the maze task studied by Paolo et al. \cite{Paolo2019} to evaluate the performance of TAXONS.

The robot always starts at the same position: at the bottom left-hand corner of the maze.
It is controlled via a Multi-Layer Perceptron (MLP) with weights and biases learned by the evolutionary process.
The MLP is used to control the robot in a closed-loop manner for $T=2000$ time-steps. 
At each time-step, the MLP takes as input the five sensor measures of the robot; and outputs the velocities commands to send to its wheels.
Each MLP has a single hidden layer of size 5.
The fitness function penalises the energy consumption: $\perf = -\sum_{t=1}^{T} \| \vec u_t \|^2$, where $\vec u_t $ is the vector of the velocity commands at time-step $t$.

The collected sensory data corresponds to a $64\times 64$ RGB image of the entire maze (including the robot) at the end of the episode; this leads to a total of 12,288 dimensions.
For reference, the hand-coded BD $\bdtask$ corresponds to the final position $(x_T, y_T)$ of the robot.
Thus, in the maze task, the sensory data $\obs$ and the hand-coded BD $\bdtask$ contain the same information: $\obs$ can be deduced from $\bdtask$ and inversely.

\subsubsection{Hexapod}

We also consider the hexapod omni-directional QD task from Cully et al. \cite{Cully2013, Cully2016}.
The purpose of this task is to learn a container of controllers for a hexapod robot such that: the different controllers bring the robot to diverse positions.
Each leg of the simulated hexapod is controlled by three actuators using sine-wave-like functions, which are parameterised by their phase, amplitude and duty cycle.
The full genotype is composed of 36 values in the interval $\left[0,1\right]$.
Additional details can be found in the work of Cully et al. \cite{Cully2014}. 
The fitness function promotes circular trajectories: $\perf = - |\alpha^{\indiv} - \alpha^{\text{circular}}|$, where $\alpha^{\indiv}$ is the final orientation of the robot, and $\alpha^{\text{circular}}$ is the desired final orientation that would be obtained if the trajectory was circular.

Each episode lasts for three seconds.
At the end of each episode, a top-down picture of the environment is taken from a camera which is placed at the vertical of the origin (Fig.~\ref{fig:all_experiments}). 
That $64\times 64$ RGB picture corresponds to the sensory data $\obs$ of the individual; the sensory data vector has thus 12{,}288 dimensions.
Similarly to previous work, the hand-coded BD $\bdtask$ corresponds to the final position $(x_T, y_T)$ of the hexapod torso \cite{Cully2013, Cully2016}.
As in the maze task, the final position $\bdtask$ of the robot can be deduced from the picture of the robot in $\obs$. 
However, in this case, the picture of the robot contains strictly more information than $\bdtask$, such as the orientation of the robot and the positions of its legs.

\subsubsection{Air-Hockey Task}

Introduced by Cully \cite{Cully2019}, the air-hockey task consists of a two-dimensional robotic arm with four Degrees of Freedom that has to learn how to push a puck to many different positions.
To do so, for the first five seconds of the experiment, the robotic arm moves to an initial configuration (Fig.~\ref{fig:all_experiments}).
Then, a puck is placed at a pre-defined fixed position in the environment.
Finally, for the following five seconds, the arm moves to another configuration.
By doing so, it may push the puck, which can move and bounce on the walls.

The genotype contains eight parameters. 
The first four parameters represent the angles of the actuators of the first configuration.
The last four parameters represent the angles of the second configuration.
The angles are defined between $-\pi$ and $\pi$.
The fitness function penalises the energy consumption: $f = -\sum_{t=1}^{T} \| \vec v^{\text{joints}}_t \|^2$, where $\vec v^{\text{joints}}_t $ is the vector of the joints velocities at time-step $t$.

The trajectory of the puck is recorded by capturing its position $(x_t, y_t)$ every $0.1$ second.
The sensory data considered in this task corresponds to the puck trajectory $\obs = \left( x_t, y_t \right)_{t\in\left[1..T\right]}$; the sensory data vector has thus 100 dimensions.
Similarly to the previous tasks, the hand-coded BD corresponds to the final position of the puck: $\bdtask = (x_T, y_T)$.
Thus, $\bdtask$ contains strictly less information compared to $\obs$, as $\obs$ contains information about the entire trajectory of the puck, including its last state.

\subsection{Metrics}

Our evaluations of the performance of each algorithm will be based on four metrics:

\subsubsection{Coverage}

\label{subsubsec:metrics-coverage}

The coverage metric estimates the percentage of coverage of the hand-coded BD space $\bdtaskspace$. 
To calculate it, each axis of the hand-coded BD space is subdivided into $40$ parts, resulting in a grid of $40\times 40$ cells.
The coverage corresponds to the percentage of cells containing at least one individual.
The hand-coded BD space considered in each task is given in Table~\ref{tab:variants_maze}.

\subsubsection{Grid Mean Fitness}

\begin{figure}[t]
  \centering
\includegraphics[width=0.45\textwidth]{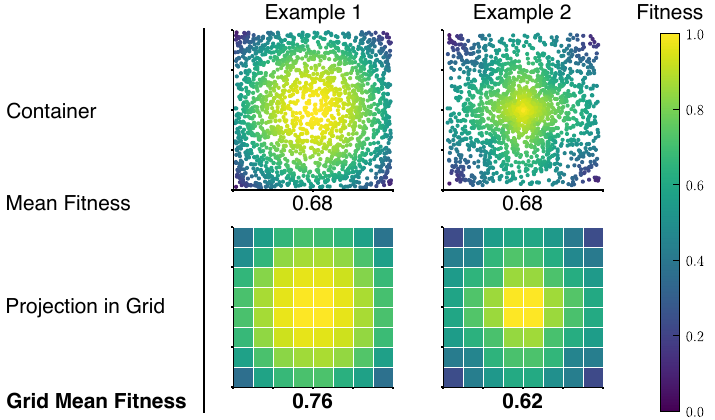}
\caption{%
Illustration of the grid mean fitness metric.
The first row presents the hand-coded BDs for two unstructured containers of 2000 individuals with different fitness distributions.
The second row indicates the approximate mean fitness of each container.
The third row shows the grid containers after having projected the hand-coded BDs into an $8\times 8$ grid, and having kept the best solution in each cell.
The last row presents an estimation of the grid mean fitness for each container.
The two containers have the same mean fitness (see second row).
However, the container on the left has a higher grid mean fitness, since its high-performing solutions are more spread in the BD space.
}
 \label{fig:illustration_grid_mean_fitness}
\end{figure}

To evaluate the overall quality of a container of individuals, we use a grid-based mean fitness of all the individuals from that container.
To compute it, all the individuals are projected in a $40\times 40$ grid of the hand-coded BD space.
Then, in each cell of the grid, we only keep the individual with the best fitness.
Finally, we average the fitnesses of those individuals to get the grid mean fitness of the container.
Using a grid from the hand-coded BD enables the metric to be independent of the distribution of solutions in the BD space, which is important for a fair comparison between algorithms using learned BD space. Furthermore, it favors containers with locally-optimal solutions (see Fig.~\ref{fig:illustration_grid_mean_fitness}).

\subsubsection{Container Size}
The container size corresponds to the number of individuals in the container.
The difference with the \textit{coverage} metric, is that the \textit{coverage} represents the percentage of grid bins with at least one individual; while the \textit{container size} represents the total amount of individuals in the container.
We use this metric to evaluate if \aurorapsat{} and \auroravat{} can return a container with approximately $\targetscC$ individuals.
This desired container size $\targetscC$ is set to $10^4$ for the Maze and Air-Hockey tasks.
For the Hexapod task, we set $\targetscC$ to $5\times 10^3$ to compensate for the higher number of QD iterations, and for the high computational cost of the simulated environment. 
These values for $\targetscC$ have the same order of magnitude as most resulting container sizes in the QD literature \cite{Cully2018, Cully2014, Vassiliades2018a, Cully2019, tarapore2016differentencodingsinfluence}.

\subsubsection{Average Container Loss per Update}

During a container update, all the individuals are removed from the container, and are re-added, taking into account the new adaptive threshold $\thresholdNov$.
If the $\thresholdNov$ has increased, some individuals may not be re-added to the container.
Those individuals are lost during the update of the container.
The average container loss per update corresponds to the total number of individuals lost divided by the number of container updates.

\subsection{Implementation Details}

The dimensionality reduction algorithm used by AURORA and TAXONS relies on a neural network Auto-Encoder (AE).
More specifically, for the Maze and the Hexapod tasks, the convolutional encoder consists of three convolutional layers of size $\left[ 4, 4, 4 \right]$, followed by two fully connected layers of size $\left[ 64, 32\right]$; the decoder consists of two fully connected layers of size $\left[ 32, 64\right]$ followed by four transposed convolutional layers of size $\left[ 4, 4, 4, 3\right]$. 
For the Air-Hockey task, the encoder is a Multi-Layer Perceptron (MLP) whose hidden layers have the sizes: $\left[32, 8\right]$; the decoder is also an MLP with hidden layers of sizes: $\left[8, 32\right]$.
Those AEs are trained using the Adam Optimiser \cite{kingma2014adam} with $\beta_1 = 0.9$ and $\beta_2 = 0.999$.
Furthermore, as explained in section~\ref{sec:discovering_unsupervised_behaviours}, the spacing between two AE updates is linearly increased. In our experiments, the first AE update occurs at iteration 10, and the other updates happen at the following iterations: 30, 60, 100, 150...
The auto-encoders are implemented and trained using the C++ API of Torch \cite{paszke2019pytorch}.
As in the experiments from from Paolo et al. \cite{Paolo2019}, by default, we use a latent space dimensionality of 10.

Our implementation relies on Sferes$_{v2}$ \cite{Mouret2010} and its integrated QD framework \cite{Cully2018}.
The maze environment uses the libfastsim simulator \cite{Mouret2012}, and the neural network controllers are evolved using the nn2 module \cite{Mouret2012}.
The hexapod environment is based on the DART simulator \cite{lee2018dart}.
Finally, the air-hockey task uses the Box2D physics engine \cite{catto2010box2d}.

Each variant was run with 20 replications.
The statistical significance of the comparisons is measured by using Wilcoxon rank-sum test. 
We use the Holm-Bonferroni method \cite{holm1979simple} to estimate the adjusted p-value thresholds for multiple comparisons.
In the following sections, all p-values will be abbreviated: $p$.
The remaining hyperparameters are provided in the appendix.
For reproducibility purposes, our source code and a singularity container \cite{kurtzer2017singularity} with pre-built experiments are available at \texttt{\url{https://github.com/adaptive-intelligent-robotics/AURORA}}.

\section{Results}

\begin{figure*}[t]
  \centering
\includegraphics[width=0.9\textwidth]{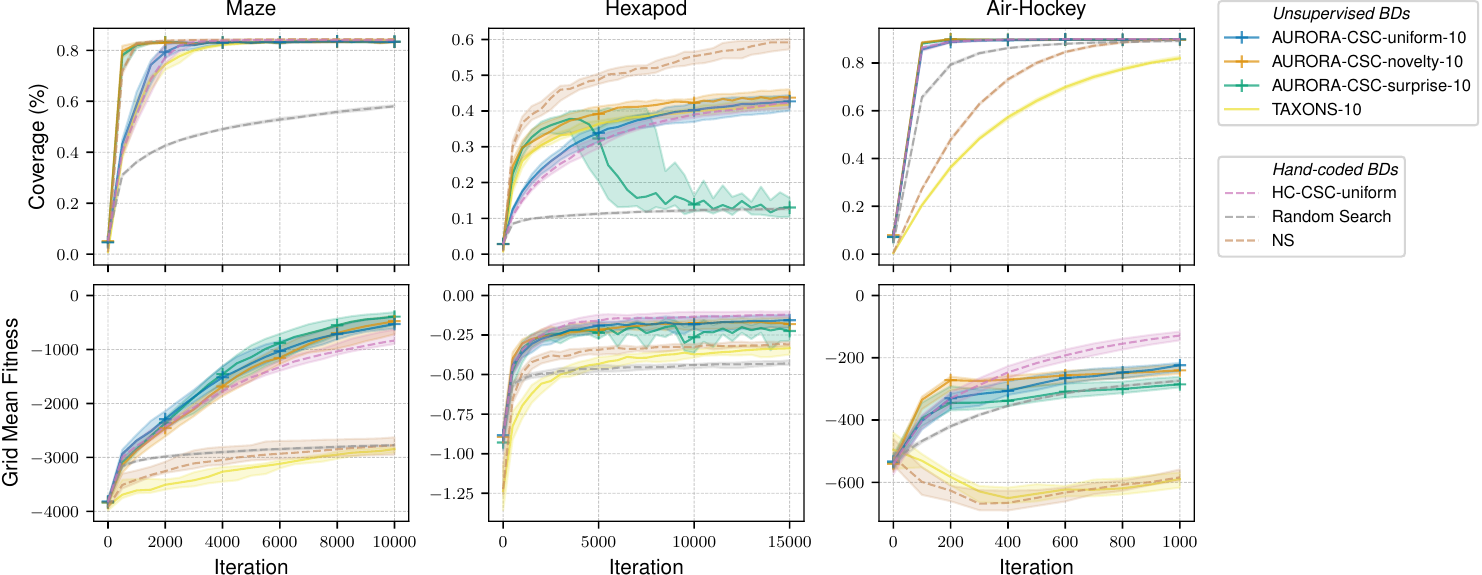}
\caption{%
Progression of the coverage and the grid mean fitness for the different variants of \aurorapsat{} and TAXONS with a latent space of dimension 10.
The results are displayed in the three environments under study (from left to right): Maze, Hexapod and Air-Hockey.
Each curve presents the medians of the results obtained from 20 replications.
The shaded areas represent the interquartile range.
An illustration of the grid mean fitness metric is provided in Fig.~\ref{fig:illustration_grid_mean_fitness}.
}
 \label{fig:comparison_variants}
\end{figure*}

\begin{figure*}[t]
  \centering
\includegraphics[width=\textwidth]{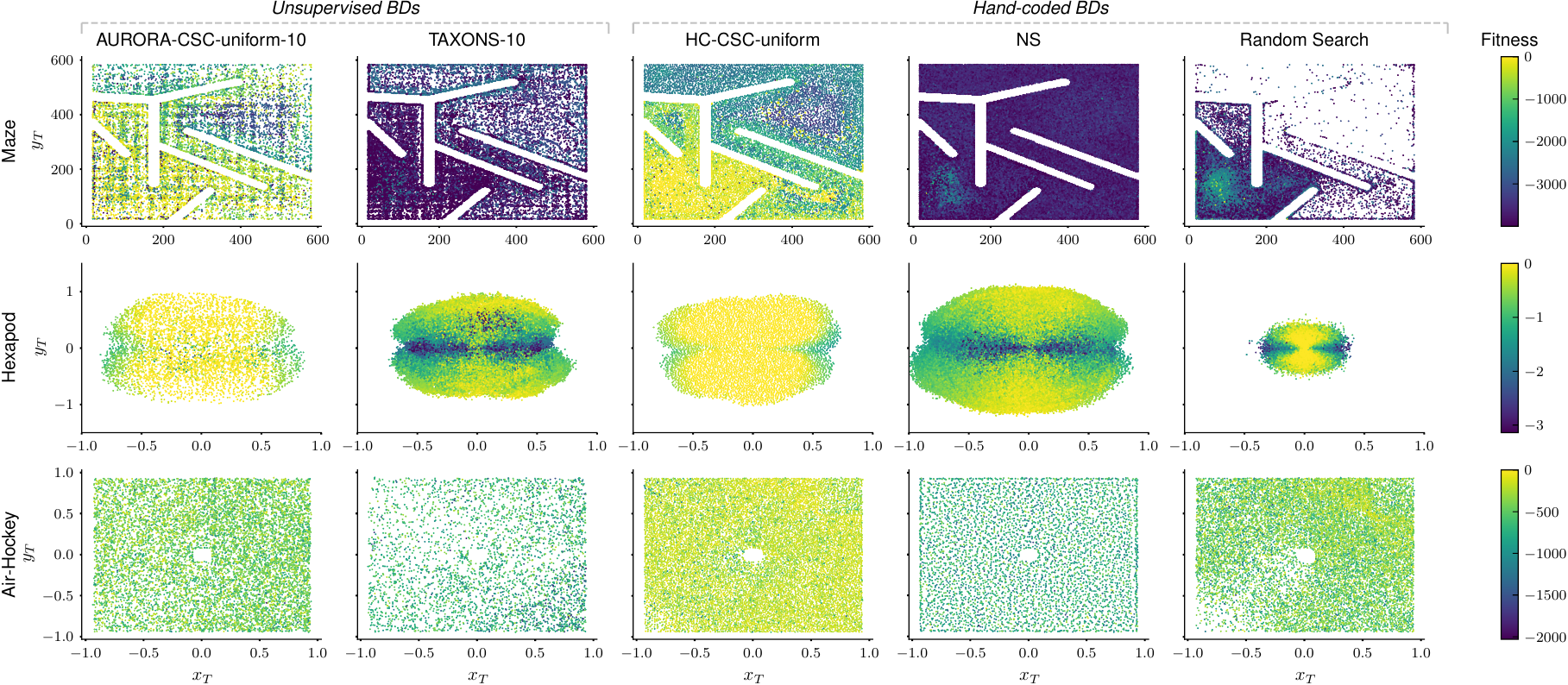}
\caption{Task Behavioural Descriptors (BDs) $\bdtask$ in containers obtained from several algorithms under study. In each sub-figure, each dot corresponds to the hand-coded BD $\bdtask$ of a single individual from the container; the color of the dot is representative of the individual's fitness.}
 \label{fig:scatter_plots_variants}
\end{figure*}

\begin{figure*}[t]
  \centering
\includegraphics[width=0.60\textwidth]{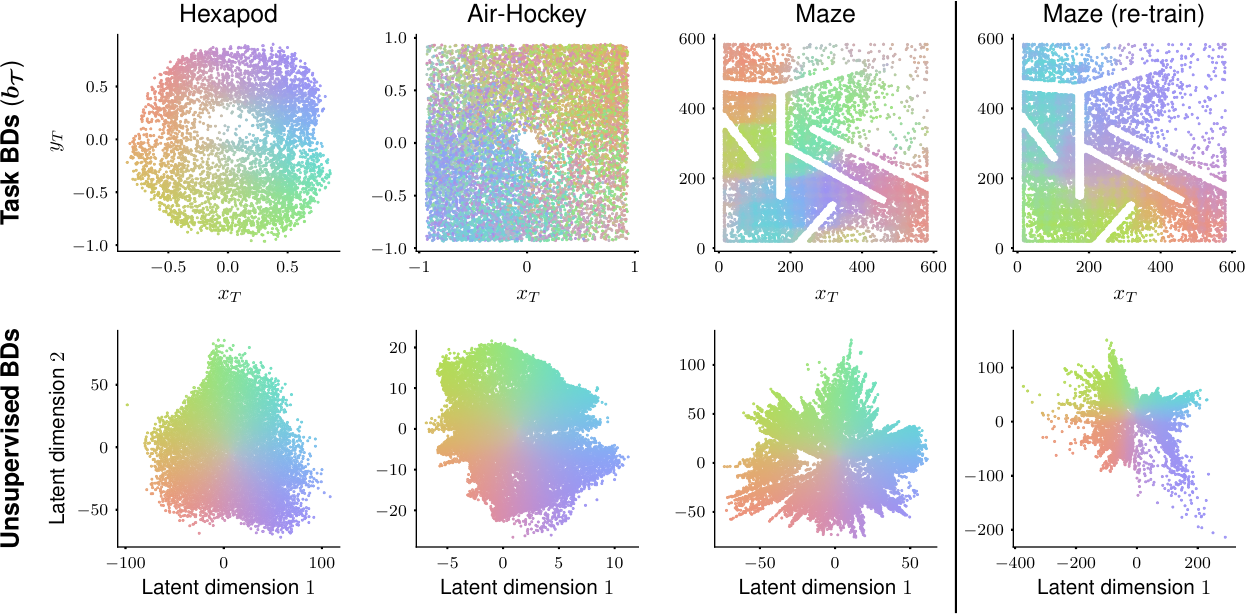}
\caption{%
(First three columns) Visualisations of the task Behavioural Descriptor (BD) space and the two-dimensional latent space learned by AURORA-CSC-uniform-2 for the three tasks under study (from left to right): the hexapod, air-hockey and maze tasks. 
The individuals were colourised via a rotation colouring, based on the positions in the latent space (bottom row). 
For each task, each individual is represented in the hand-coded BD space and in the latent space using the same colour.
(Last column) Same container of individuals as the one shown in the third column for the maze task. 
However, the latent space results from a reset and re-training of the encoder; whereas the results shown in the first three columns result from incremental training procedures of the encoder.
}
 \label{fig:meaningful_latent_space}
\end{figure*}

This section is structured around three parts:

First, we study to which extent AURORA inherits the optimization capabilities of standard Quality-Diversity (QD) algorithms.
We show that \aurorapsat{} fulfill its goal of finding a container of diverse high-performing abilities without requiring hand-coded Behavioral Descriptors (BDs).

Second, we study the features present in the BDs learned by AURORA, and how those features influence the diversity in the container. 
As explained in section~\ref{sec:tasks_description}, the problem setting differs between AURORA and its baselines: AURORA learns its BDs from raw sensory data, whereas the baselines rely on less-informative hand-coded BDs.
On the one hand, we study the relationship between the unsupervised BDs and the ground truth states of the robot.
On the other hand, we analyse the capability of AURORA to exploit sensory data features that are not present in the hand-coded BDs.

Finally, we study the robustness of AURORA to some of its hyperparameters. 
To do so, we first evaluate the influence of the BD dimensionality.
Then, we analyse how the container and encoder update periods affect the optimisation procedure.

\subsection{Performance of \aurorapsat{}}

\subsubsection{Learning Diverse Abilities}

In this section, we show that AURORA can discover diverse behaviours for an agent.
Fig.~\ref{fig:comparison_variants} shows that even though it does not have access to the hand-coded BDs, AURORA reaches a decent coverage in the hand-coded BD space.
In the three tasks under study, AURORA presents a faster increase in coverage than TAXONS.
In particular, in the air-hockey task, AURORA reaches a coverage of the hand-coded BD space which is significantly higher ($p < 3\times 10^{-6}$).

Despite the differences between the three tasks, AURORA-CSC-uniform always achieves a coverage close to the upper bound baseline HC-CSC-uniform.
The similarity between those coverage performances results from the fact that, in all tasks, the hand-coded BDs can be deduced from the sensory data.
Interestingly, the Novelty Search (NS) baseline obtains a substantially higher coverage in the hexapod task, compared to all the other variants under study ($p < 2\times 10^{-6}$).

We can also observe that, as for standard QD algorithms \cite{Cully2018}, the selector type has an impact on the exploration efficiency of AURORA.
For example, in the maze and hexapod tasks, AURORA obtains a significantly higher coverage at iteration 1000 when using a novelty-based and surprise-based selector instead of a uniform selector ($p<2\times 10^{-6}$).
In the maze and air-hockey tasks, the surprise-based and novelty-based variants of AURORA reach a similar coverage.
However, in the hexapod task, the coverage of the surprise-based variant drops significantly after $5000$ iterations ($p < 4\times 10^{-2}$ at iteration 15000).
We hypothesise that this decline is caused by a decrease in the performance of the auto-encoder.
Such a decrease in performance may be due to the surprised-based selector, which promotes individuals having the worst reconstruction from the auto-encoder.
In this case, the container may contain a consequent amount of outliers, which may affect negatively the training of the auto-encoder.

\subsubsection{Learning High-Performing Abilities}

The grid mean fitness also seems to be related to the discrepancy between the hand-coded BD and the sensory data.
For example, in the maze and hexapod tasks, the grid mean fitness obtained by AURORA is close to the one obtained with its upper-bound variant HC-CSC-uniform (Fig.~\ref{fig:comparison_variants}).
In those tasks, the containers learned by AURORA and the hand-coded baselines exhibit the same patterns in the fitness distribution (Fig.~\ref{fig:scatter_plots_variants}). 
When there is a high discrepancy between the hand-coded BD and the sensory data, as in the air-hockey task, the grid-based mean fitness achieved by the hand-coded baseline gets better compared to AURORA (Figs.~\ref{fig:comparison_variants} and~\ref{fig:scatter_plots_variants}, $p<2\times 10^{-6}$). 
However, we will see in the next subsection that this is compensated by an increased diversity when the analysis is not restricted to hand-coded BDs.

Contrary to AURORA and its hand-coded variants, NS and TAXONS are exploration-only algorithms.
They do not intend to maximise the fitness scores of their solutions.
That explains the low grid-based mean fitness compared to AURORA (Fig.~\ref{fig:comparison_variants}, $p < 2\times 10^{-2}$), and the low fitness scores of the individuals present in their containers (Fig.~\ref{fig:scatter_plots_variants}).

\begin{figure*}[t]
  \centering
\includegraphics[width=\textwidth]{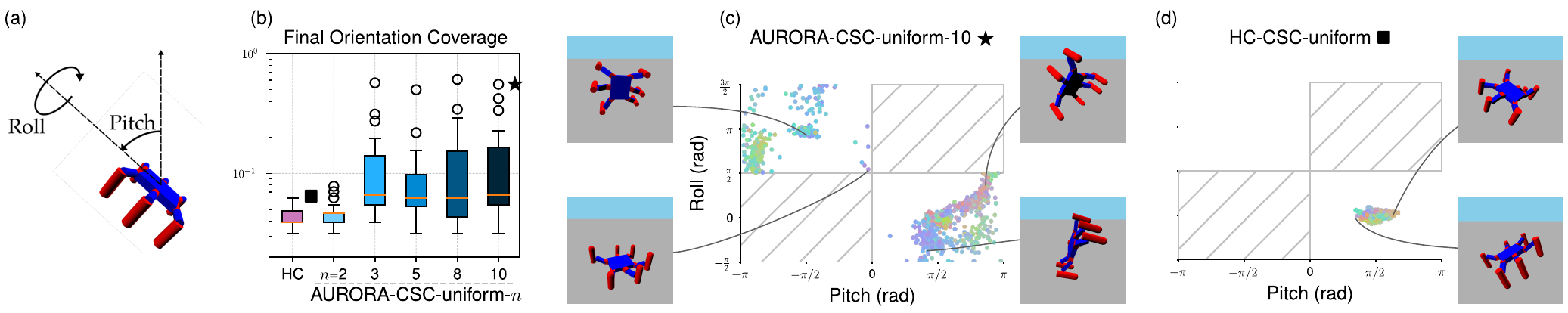}
\caption{(a) Illustration of the pitch and roll Euler angles for the hexapod robot.
(b) Final orientation coverage obtained for the Hand-Coded variant (HC-CSC-uniform), and AURORA-CSC-uniform-$n$ for various BD dimensionalities $n$.
The star $\bigstar$ and the square $\blacksquare$ each represent the status of the best replication at the last iteration.
(c) and (d): Combination of pitch-roll angles present in the container for the $\bigstar$ and the $\blacksquare$ replications.
Each marker is colourised based on the final position $(x_T, y_T)$ of the individual.
Some pitch-roll combinations are equivalent (for example: $\left(\frac{\pi}{2}, 0\right)$ and $\left(-\frac{\pi}{2}, 0\right)$).
To avoid showing several times the same individuals, we define a specific space for pitch-roll angles.
This defined space corresponds to the areas that are not hashed.
}
 \label{fig:hexapod_diversity}
\end{figure*}

\subsection{Study of the Behavioural Descriptors Learned by AURORA}

\subsubsection{Correlation Between Unsupervised Behavioural Descriptors and Ground Truth}

Even though AURORA does not have direct access to the ground truth, the unsupervised BD space remains meaningful with respect to the hand-coded BDs.
Indeed, the two-dimensional latent space learned by AURORA-CSC exhibit some correlation with the hand-coded BD space (Fig.~\ref{fig:meaningful_latent_space}). 
For instance, in the hexapod task, the corners appear distinctly in the hand-coded BD space and in the latent space.
In the air-hockey task, the unsupervised BDs and the hand-coded BDs still appear to be correlated.
However, several individuals having different unsupervised BDs have similar hand-coded BDs.
Knowing that unsupervised BDs are learned from full puck trajectories, and that the hand-coded BD corresponds to the final position of the puck, this suggests that AURORA learns several puck trajectories arriving in the same area.
In the maze task, even though the correlation is still apparent, several individuals that are far from each other in the latent space, are very close in the hand-coded BD space (Fig.~\ref{fig:meaningful_latent_space}).
This BD space entanglement is interesting, as there is no discrepancy between the information contained in the sensory data and the hand-coded BDs.

To investigate the reasons for that BD space entanglement, we trained from scratch an encoder on the same final container for the maze task.
This approach differs from the encoder training in AURORA, which is incremental.
After having re-trained the encoder, the resulting latent space reveals more coherence than the one initially obtained by AURORA (Fig.~\ref{fig:meaningful_latent_space}).
This suggests that, when the latent space has two dimensions only, and when the space is progressively explored irregularly, it is complicated for the encoder to incrementally update the manifold of the unsupervised BDs in a meaningful way.

\subsubsection{Ability to Learn Novel Behaviours with respect to Additional Features}

In this section, we intend to show that AURORA not only learns a meaningful BD structure, but is also capable of improving the container diversity with respect to more features than the hand-coded BD.
We study this extensive diversity on the hexapod and air-hockey tasks.
In those two tasks, the sensory data contains strictly more information related to the agent than the hand-coded BD.

For the hexapod task, we evaluate the diversity of final orientations of the hexapod robot.
In particular, we assess the diversity of final pitch and roll Euler angles of the hexapod (see illustration in Fig.~\ref{fig:hexapod_diversity}).
We do not consider the yaw angles as the fitness function encourages solutions to reach a target angle. 
For the analysis, we discretize the pitch/roll angles space into a $16\times 16$ grid; and we define the orientation coverage as the percentage of grid cells filled.
We compare AURORA-CSC-uniform with various latent space dimensionalities, to its equivalent hand-coded variant HC-CSC-uniform.
Among those variants, our results show that AURORA achieves the widest orientation coverage when the dimensionality of the unsupervised BD space exceeds two (Fig.~\ref{fig:hexapod_diversity}, $p < 2\times 10^{-3}$).
For instance, some replications of AURORA find how to make the hexapod roll over on its back.

\begin{figure}[t]
  \centering
\includegraphics[width=0.48\textwidth]{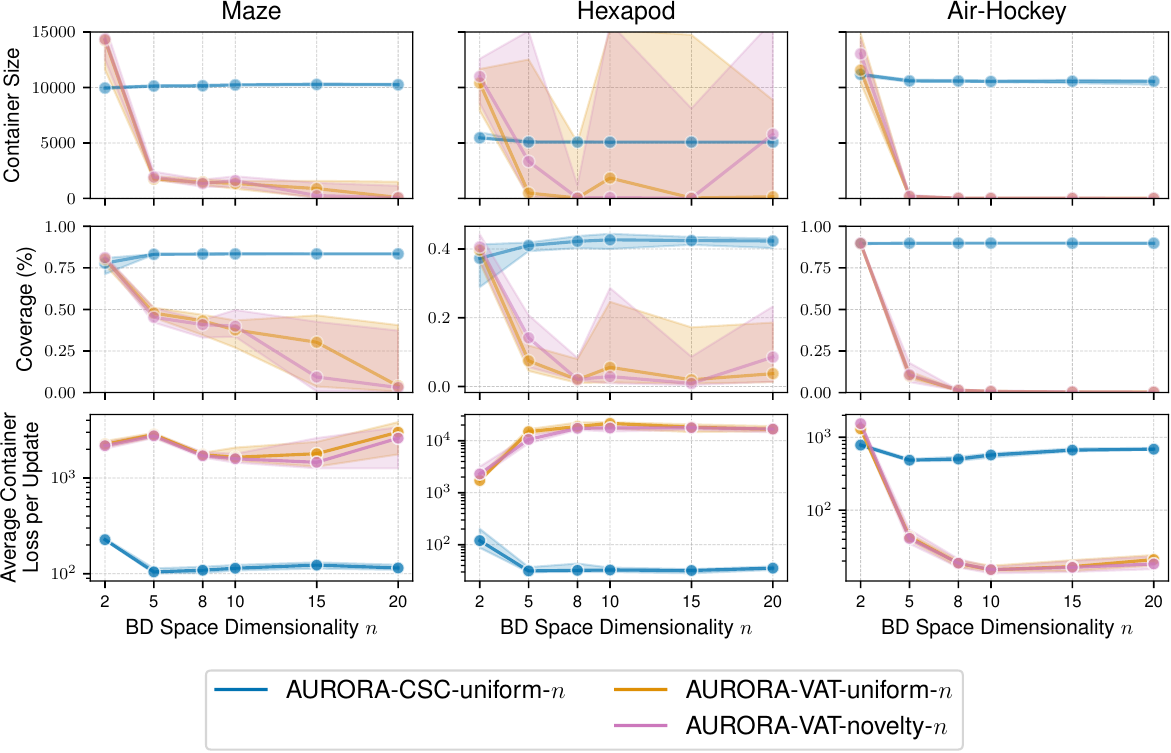}
\caption{Influence of the unsupervised Behavioural Descriptor (BD) space dimensionality on \aurorapsat{} and \auroravat{}.
The first row shows the evolution of the container size achieved at the end of the experiment for the three environments under study (note that $\targetscC=5\times 10^3$ for the hexapod task; and $\targetscC=10^4$ for the maze and air-hockey tasks). The second row presents the coverage of the hand-coded BD space achieved at the end of the experiment. The third row shows the mean population loss per container update, depending on the dimensionality of the latent space.}
 \label{fig:influence_latent_dim}
\end{figure}

As expected, when the dimensionality of the unsupervised BD space is low, AURORA-CSC takes fewer features into account from the sensory data.
For example, the two-dimensional version of AURORA presents a lower orientation coverage than the other multi-dimensional variants (Fig.~\ref{fig:hexapod_diversity}, $p < 5 \times 10^{-3}$).
This difference of performance is less apparent with the position coverage: for all tested latent space dimensionalities, AURORA learns a set of behaviours ending at diverse positions (Fig.~\ref{fig:influence_latent_dim}).

\begin{figure}[t]
  \centering
\includegraphics[width=0.48\textwidth]{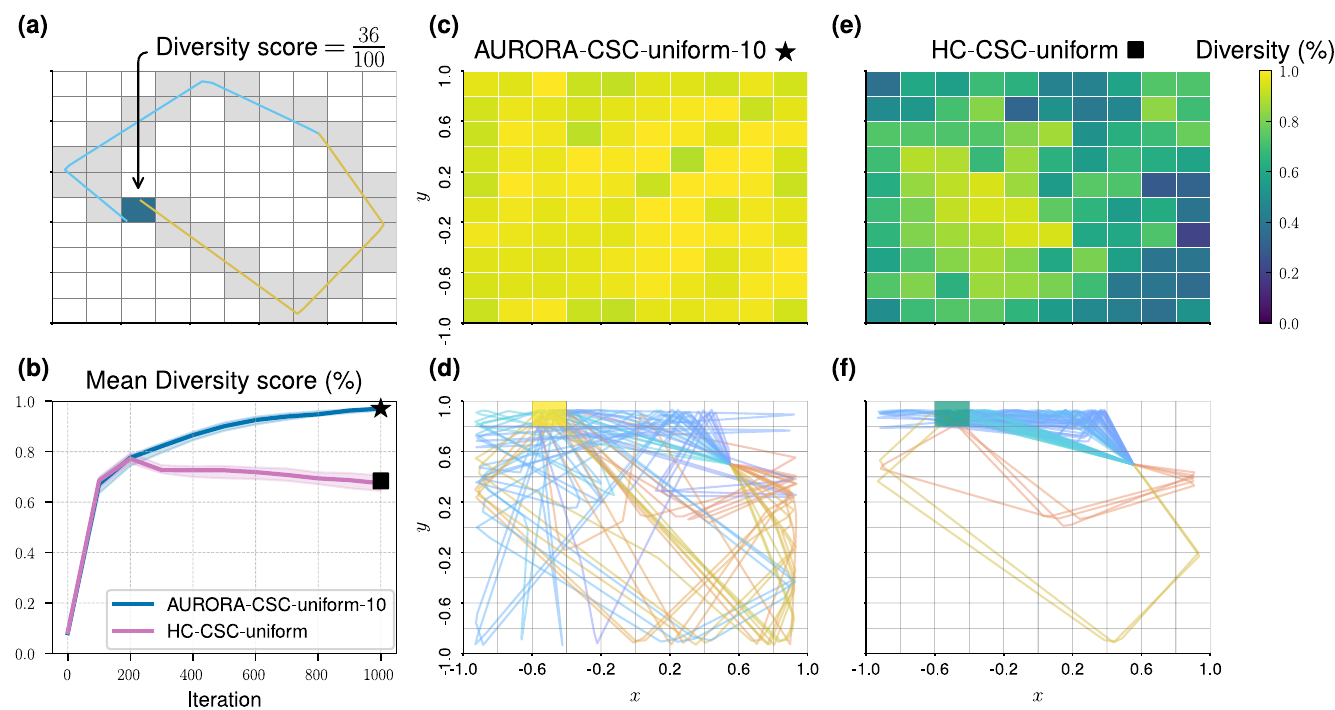}
\caption{%
(a) Illustration of the diversity score of one grid cell in the air-hockey task.
Each coloured curved line represents a puck trajectory.
(b) Progression of the mean diversity score for AURORA and the hand-coded baseline HC-CSC-uniform. 
The mean is calculated from all the diversity scores of all the grid cells.
The star $\bigstar$ and the square $\blacksquare$ represent the status of one replication at iteration 1000.
(c) and (e): Diversity scores per grid cell for the $\bigstar$ and the $\blacksquare$ replications.
(d) and (f): All puck trajectories ending in the grid cell $\left[ -0.6, -0.4 \right]\times \left[ 0.8, 1 \right]$.
Each trajectory is colourised based on its initial angle.
}
 \label{fig:air_hockey_diversity}
\end{figure}

\smallskip

In the air-hockey task, we estimate the diversity of puck trajectories following the approach used by Cully \cite{Cully2019}.
The space of final puck positions is discretised into a $10\times 10$ grid, and each cell of the grid is attributed a diversity score.
To compute the diversity score of one grid cell, we consider all the puck trajectories ending in that cell, and we evaluate the percentage of grid cells crossed by those trajectories (see Fig.~\ref{fig:air_hockey_diversity}a).
Similarly to the hexapod task, we compare the performance of AURORA-CSC-uniform and its hand-coded equivalent.
The results indicate that AURORA achieves the highest mean diversity score ($p < 2\times 10^{-6}$).
Furthermore, most grid cells seem to have a better diversity score in AURORA (Fig.~\ref{fig:air_hockey_diversity}).
Consequently, for each grid cell, the puck trajectories ending in that cell appear to be more diverse in AURORA.

\smallskip

In the end, the experimental results show that the behavioural descriptors learned by AURORA consider multiple features extracted from the sensory data. 
If the sensory data contains strictly more information than the hand-coded BDs, then AURORA may learn a more general diversity measure compared to its hand-coded baselines.

\subsection{Hyperparameters Sensitivity}

\subsubsection{Robustness of \aurorapsat{} and \auroravat{} to Latent Space Dimensionality}

In AURORA, the dimensionality of the unsupervised BD space used can be adjusted depending on the number of details we want to grasp in the sensory data.
Therefore, AURORA needs to be robust to the number of dimensions in the BD space.
In this section, \aurorapsat{} is compared with its \auroravat{} variant, which was suggested by Cully \cite{Cully2019}.
We consider both variants with a uniform selector.
To ensure that the results obtained by \auroravat{} are not due to its selector, we also consider for comparison \auroravat{} with a novelty-based selector (this selector has been shown to improve the performance of \aurorapsat{} in Fig.~\ref{fig:comparison_variants}).

With \aurorapsat{}, the container size and coverage remain mostly stable for all dimensions (Fig.~\ref{fig:influence_latent_dim}).
It can be noticed that the container size remains close to the desired value $\targetscC$.
On the contrary, with both variants of \auroravat{}, the container size gets unstable when the dimensionality exceeds four.
For such dimensionalities, the final coverage achieved using the VAT method is systematically lower than the CSC variant ($p<4\times 10^{-4}$).
The way \auroravat{} operates in high-dimensional spaces can be divided into two categories.

On the one hand, in the air-hockey task, the average container loss is low compared to the CSC variant (Fig.~\ref{fig:influence_latent_dim}, $p<5\times 10^{-5}$), and the container size is close to zero when the dimensionality of the latent space exceeds four.
That low container size indicates that it is difficult for new individuals to be added to the container.
This suggests that the estimations of the distance threshold $\thresholdNov$ made by the VAT method are too high.
On the other hand, in the maze and hexapod tasks, the average container loss is high compared to \aurorapsat{} (Fig.~\ref{fig:influence_latent_dim}, $p<10^{-5}$).
Consequently, the container size is more unstable in \auroravat{}: more individuals are added to the container, and more individuals are deleted during container updates.
This suggests that \auroravat{} alternates between two states: (1) a state where it under-estimates $\thresholdNov$, making it easy for new individuals to be added to the container; and (2) a state where it over-estimates $\thresholdNov$, which removes all the individuals previously added when the container is updated.

Thus, \aurorapsat{} is more robust to the BD space dimensionality compared to its VAT variant.
In particular, it achieves its purpose of returning a container with approximately $\targetscC$ individuals.

\subsubsection{Influence of Container and Encoder Update Periods}

\begin{figure}[t]
  \centering
\includegraphics[width=0.48\textwidth]{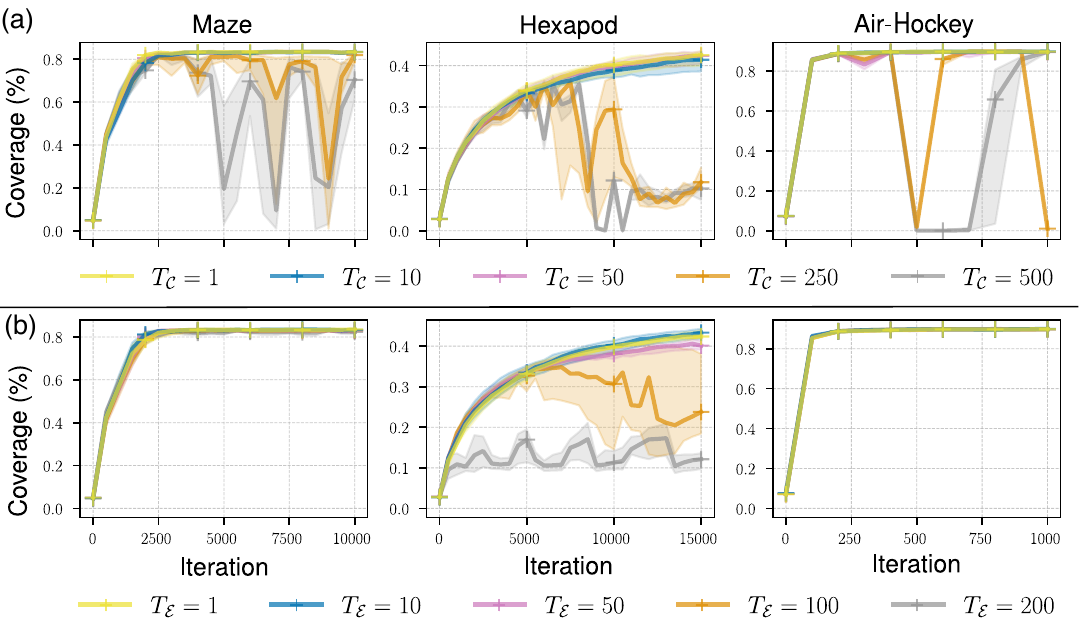}
\caption{(a) Evolution of coverage for various container update periods $T_\cC$. 
More precisely, $T_\cC$ represents the amount of iterations between two container updates; the lower it is, the more frequent are the container updates.
(b) Evolution of coverage for various values of $T_\encoder$. 
The interval between two encoder updates increases linearly: it follows an arithmetic sequence; $T_\encoder$ represents the common difference of that arithmetic sequence.
For example, if $T_\encoder=100$, then the encoder updates occur at iterations: 100, 300, 600...
}
 \label{fig:encoder_and_container_update}
\end{figure}

Finally, we study the effect of the container update period $T_\cC$, and of the update period of the encoder.
More specifically, we evaluate how those hyperparameters influence the evolution of coverage for AURORA-CSC-uniform-10.

We study the evolution of the coverage obtained for various values of container update periods $T_\cC$.
Figure~\ref{fig:encoder_and_container_update}a shows that when $T_\cC$ is too high, there are significant declines in coverage during the optimisation process.
Those declines notably lead to a lower coverage when $T_\cC = 500$ ($p<3\times 10^{-4}$ if compared to the $T_\cC=10$ variant).
Those declines are caused by two mechanisms:
(1) in \aurorapsat{}, the value of $\thresholdNov$ is updated at each iteration;
and (2), following update rule~\ref{eq:update-l}, if the container size is bigger than $\targetscC$, then the distance threshold $\thresholdNov$ increases.
Thus, if $T_\cC$ is too high and if the container size is larger than $\targetscC$, then  $\thresholdNov$ may increase excessively; and the higher $\thresholdNov$ is, the more individuals are deleted from the container when it is updated.
That is why the deletions of individuals may be excessive when there are too many iterations between two container updates.

In the previous experiments, the duration between two encoder updates increases linearly over iterations.
In other words, the interval between two encoder updates is an arithmetic sequence.
We write $T_\encoder$ the common difference of this arithmetic sequence.
For instance, $T_\encoder$ is equal to 10 in all our previous experiments (as explained in~\ref{sec:encoder_updafte_phase}); and if $T_\encoder=20$, then encoder updates occur at the following iterations: 20, 60, 120, 200...
As presented in Figure~\ref{fig:encoder_and_container_update}b, an excessively high $T_\encoder$ may lead to a significant decline in coverage.
This is particularly noticeable in the Hexapod task when $T_\encoder$ is superior to 100 ($p<2\times 10^{-2}$).
If the encoder is not updated frequently enough, then the encoder may not capture the relevant features in the varying sensory data.
For example, if the entire container is modified between two encoder updates, then the encoder dataset changes considerably.
In that case, each encoder update may modify significantly the behavioral descriptors present in the container, leading to instability in the optimisation.

If $T_\cC$ or $T_\encoder$ are excessively high, then the QD algorithm may obtain a worse coverage.
However, if those values are very low, then too many container and encoder updates will be performed.
As those updates are time-consuming, it is relevant to find a trade-off when choosing the values of $T_\cC$ and $T_\encoder$.

\section{Conclusion and Discussion}

In this work, we have presented AURORA, a Quality-Diversity (QD) algorithm that learns by itself its behavioural descriptors in an unsupervised manner.  
Even though AURORA does have direct access to a hand-coded BD, it succeeds to build a container of diverse behaviours.
We also show that AURORA can be used, like any QD algorithm, to optimise the quality of the discovered behaviours.
The performance of AURORA depends on the discrepancy between the sensory data collected by each individual and the hand-coded BDs.
In case the sensory data contains more information than the hand-coded BDs, the unsupervised BD space learned by AURORA takes more features into account than the hand-coded BD space.

To maintain the container size around a targeted value, we introduced the Container Size Control (CSC) technique.
AURORA-CSC has been proven to be robust to the dimensionality of the learned BD space.
For the moment, the CSC method relies on a proportional control loop.
It would be relevant to try to improve the CSC method by using more elaborated control techniques, such as a proportional–integral–derivative controller for example.
Moreover, the CSC method supposes that there is no sudden significant alteration of the whole BD space.
Such strong alteration may happen during an encoder update, especially if the encoder parameters are reset.
It might be interesting to study how to adapt the CSC method to take into account such changes in the BD space.

All the tasks under study in this paper are locomotion tasks in simulation.
One direct follow-up of this work is to evaluate the performance of AURORA in real-world or grasping tasks, as done by Laversanne-Finot et al. \cite{laversanne2020intrinsically}.
However, real-world locomotion scenarios add several layers of complexity to the problem. 
In such scenarios, the task does not reset, and the environment may be only partially observable.
Those layers of complexity represent interesting future research directions.
It will also be interesting to investigate how AURORA could be applied to other domains than robotics, such as procedural content generation for games \cite{Liapis2013} and engineering design \cite{gaier2017dataefficient}.

In the three tasks considered in this paper, the hand-coded BD $\bdtask$ corresponds to the final $(x,y)$ position of the robot.
We may also want to consider the behaviour of the robot over its entire trajectory.
In the Hexapod Straight Walking task from Cully et al. \cite{Cully2014, Cully2018}, the hand-coded BD is the proportion of time each leg is in contact with the ground. 
In that case, the BD is a function of the sensory data collected over its full trajectory.
The performance of AURORA in such a task would depend on the considered sensory data; it could be a sequence of measures from all joints collected over an episode, or a sequence of only feet sensors data.
In those cases, AURORA would probably benefit from more suitable encoding techniques.
For instance, encoding structures such as LSTM-based AEs \cite{sutskever2014sequence} could be used to assimilate the temporal aspect of the sensory data. 
Such structures would be particularly useful for encoding sequences of different sensory data streams like joint positions and velocities, feet contact sensors, or videos from external and embedded cameras.   
The study of other types of encoding techniques that regularize the latent space, such as Variational AEs \cite{kingma2013auto}, would also be of interest.

\appendix[Quality-Diversity Hyperparameters]

\label{appendix:other_hyperparameters}

Table~\ref{tab:hyperparameters} presents all the hyperparameters considered.
For each task, the hyperparameters are the same for all algorithms.

\begin{table}[htbp]
    \centering
        \caption{Quality-Diversity Hyperparameters}

    \begin{threeparttable}
    \begin{tabular}{l | l l l}
        \toprule

         & Maze & Hexapod & Air-Hockey \\
\midrule
            	\addlinespace

        Number of iterations $I$ & $10\times 10^{3}$ & $15\times 10^{3}$& $1\times 10^{3}$\\
        Batch Size & $128$ & $128$ & $128$ \\
        Mutation type & polynomial \cite{deb1999nichedpolynomialmutation} & polynomial & polynomial \\
        Mutation rate & $5\%$ & $5\%$ & $15\%$ \\
        $\eta_m$ \cite{deb1999nichedpolynomialmutation} & $10$ & $10$ & $10$ \\
        Cross-over & disabled & disabled & disabled \\

            	\addlinespace

\textit{Container Parameters} & & & \\
        $k$\tnote{1} & $15$ & $15$ & $15$ \\
        $\epsilon$\tnote{2} & $0.1$ & $0.1$ & $0.1$ \\
        $\targetscC{}$ & $10 \times 10^3$ & $5 \times 10^3$ & $10 \times 10^3$\\
        $\constantCSC$ & $5\times 10^{-6}$ & $5\times 10^{-6}$& $5\times 10^{-6}$\\
        $\constantVAT$ & $25$ & $18$ & $18$\\
        $T_\cC$ & $10$ & $10$ & $10$ \\

            	\addlinespace

\textit{TAXONS}  \cite{Paolo2019} & & & \\
        $Q$ & $5$ & $5$ & $5$ \\
        \bottomrule
    \end{tabular}
    
        \begin{tablenotes}
         \item[1] $k$ refers to the number of nearest-neighbors considered to compute the novelty score of an individual.
         \item[2] $\epsilon$ is a parameter of the "exclusive $\epsilon$-dominance" criterion for replacing an individual in the unstructured archive \cite{Cully2018}.
     \end{tablenotes}
    \end{threeparttable}
    \label{tab:hyperparameters}
\end{table}

%




\section*{Acknowledgment}

This work was supported by the Engineering and Physical Sciences Research Council (EPSRC) grant EP/V006673/1 project REcoVER.




%



\bibliographystyle{IEEEtran}
\bibliography{bibliography/bibliography}

\vfill







\end{document}